\documentclass[letterpaper, 10 pt, conference]{ieeeconf}
\IEEEoverridecommandlockouts
\usepackage[noadjust]{cite}
\usepackage{amsmath,amssymb,amsfonts}
\usepackage{algorithmic}
\usepackage{graphicx}
\usepackage{textcomp}
\usepackage{xcolor}
\usepackage{stmaryrd} %for integer brackets
\usepackage{caption}
\usepackage{subcaption}
\usepackage{algorithm}
\usepackage{algorithmic}
\usepackage{flushend} %to equalize columns at the end
\usepackage{tikz}

\newcommand\copyrighttext{%
\footnotesize \textcopyright 2021 IEEE. Personal use of this material is permitted. Permission from IEEE must be
obtained for all other uses, in any current or future media, including
reprinting/republishing this material for advertising or promotional purposes, creating new
collective works, for resale or redistribution to servers or lists, or reuse of any copyrighted
component of this work in other works.
}

\newcommand\copyrightnotice{%
\begin{tikzpicture}[remember picture,overlay]
\node[anchor=south,yshift=10pt] at (current page.south) {\fbox{\parbox{\dimexpr\textwidth-\fboxsep-\fboxrule\relax}{\copyrighttext}}};
\end{tikzpicture}%
}
\DeclareMathOperator*{\argmax}{arg\,max}

\def\BibTeX{{\rm B\kern-.05em{\sc i\kern-.025em b}\kern-.08em
    T\kern-.1667em\lower.7ex\hbox{E}\kern-.125emX}}
    
\begin{document}

\title{Plane and Sample: Maximizing Information about Autonomous Vehicle Performance using Submodular Optimization
	%{\footnotesize \textsuperscript{*}Note: Sub-titles are not captured in Xplore and
	%should not be used}
	%\thanks{All authors are with Motional, Boston, MA}
}

\author{Anne Collin$^{1}$, Amitai Y. Bin-Nun$^{1}$, Radboud Duintjer Tebbens$^{1}$ % 
	% <-this % stops a space
	\thanks{$^{1}$ Motional. Boston, MA.}%
}

\maketitle

\copyrightnotice

\begin{abstract}
	As autonomous vehicles (AVs) take on growing Operational Design Domains (ODDs), they need to go through a systematic, transparent, and scalable evaluation process to demonstrate their benefits to society. Current scenario sampling techniques for AV performance evaluation usually focus on a specific functionality, such as lane changing, and do not accommodate a transfer of information about an AV system from one ODD to the next. In this paper, we reformulate the scenario sampling problem across ODDs and functionalities as a submodular optimization problem. To do so, we abstract AV performance as a Bayesian Hierarchical Model, which we use to infer information gained by revealing performance in new scenarios. We propose the information gain as a measure of scenario relevance and evaluation progress. Furthermore, we leverage the submodularity, or diminishing returns, property of the information gain not only to find a near-optimal scenario set, but also to propose a stopping criterion for an AV performance evaluation campaign. We find that we only need to explore about 7.5\% of the scenario space to meet this criterion, a 23\% improvement over Latin Hypercube Sampling.
\end{abstract}

\section{Introduction}

As the autonomous vehicle (AV) industry is shifting its focus from prototyping vehicles to delivering safe products to society \cite{APTIV2019}, the need for a systematic performance evaluation procedure is growing dramatically. Furthermore, AV's Operational Design Domains (ODDs), defined as an intersection of many factors such as road infrastructure and weather \cite{Koopman2019}, are bound to grow, calling for scalable evaluation methods.
In addition to statistical methods examining rates of specific events per millions of miles driven, scenario-based evaluation is an important method to test the AV in specific driving situations \cite{InternationalOrganizationforStandardization2019, APTIV2019, Motional2021}.
Although the number of driving scenarios may be theoretically infinite, the space of possible scenarios can be discretized into a finite number. Nevertheless, reasonable discretization schemes are likely to create scenario spaces large enough that strategic search options are necessary to efficiently evaluate the performance of the AV.

Systematic sampling methods for the validation of the entire AV system either try to form a set of scenarios representative of real world event rates \cite{DeGelder2019, Hauer2019,Zhao2017b}, or search for scenarios in which weaknesses or abrupt performance changes in the AV are apparent \cite{Gangopadhyay2019,Mullins2017,Corso2019}. These methods typically center on a specific AV functionality, such as lane changing or vehicle following.
The former method is challenging because of the difficulty in assessing the frequency of specific events in the AV's ODD. The latter method converges to scenarios in which systems violate a pre-defined criterion, but does not allow the estimation of the violation rate of that criterion. In both methods, even a small change of ODD would require a re-estimation of frequencies or analysis of new functionalities.

In this paper, we propose a hierarchical statistical representation of AV performance, as well as a method to determine a scenario set to evaluate this performance across ODDs. The hierarchical structure supports the reuse of information from one ODD to the next, and mitigates the sensitivity of the performance estimation to event frequencies, as the method does not use them directly. We offer a criterion for assessing the value of testing on a specific scenario, and our sampling method provides guarantees of near-optimality with respect to this criterion. This is, to the authors' knowledge, the first study that offers such a guarantee with a scenario selection method.

\begin{figure}[t]
	\vspace{0.2cm}
	\centerline{\includegraphics[width=0.98\columnwidth]{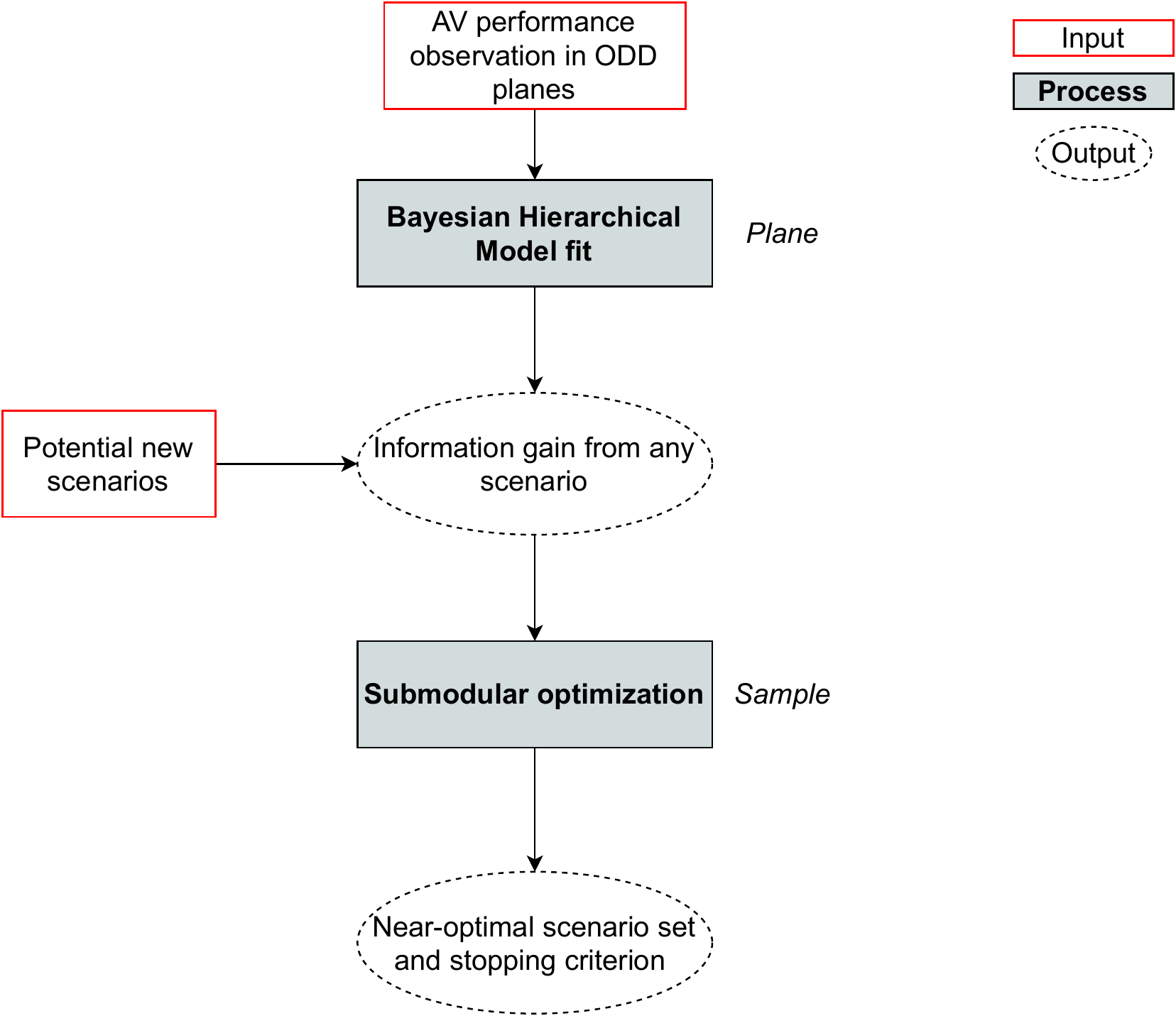}}
	\caption{Overview of the methodology presented in this paper. The Bayesian Hierarchical Model is used to compute the information gain, which is the objective function of the optimization problem used for sampling.}
	\label{fig:process}
\end{figure}

The method comprises the following steps (summarized in Fig.~\ref{fig:process}): first, we model AV performance as a Bayesian Hierarchical Model. The emergence of formal rules of expected AV behavior, and attached violation metrics, facilitates the quantitative assessment of AV performance in any scenario \cite{Censi2019, Mehdipour2019, Maierhofer2020}.
We use the factors that compose an ODD, such as town of operation, as levels in the hierarchy. This representation leverages ODD factors as units of analysis, each with just a few observations. Indeed, it becomes difficult to observe scenarios corresponding to an ODD  where all the factors are fixed, for example blue hospital signs with a specific road curvature with a high rain density and low communication reliability. The number of qualifying observations decreases with the amount of fixed factors. We call this part of the method \textit{plane}, because each level of the hierarchy effectively constitutes a hyperplane of the previous level. %and because it's a cool name hehe 

Second, we perform inference on AV performance through this Bayesian Hierarchical Model. This allows the computation of information gain on the system provided by each new scenario, which is the metric we propose to evaluate the relevance of a new scenario and of a scenario set. The Bayesian Hierarchical Model provides conditional independence properties between AV performance in different scenario space hyperplanes, causing the information gain to be submodular. This means that the information gain has a diminishing returns property; as we add more scenarios, the increase in information becomes marginal. Thus, we leverage the greedy algorithm for submodular optimization to provide a scenario set with a guarantee of a near maximal level of information gain (\textit{sample}). Our stopping criterion is when the information gain does not grow anymore given a statistical confidence level. We illustrate the method with performance obtained by simulating an AV on routes from the CARLA challenge \cite{CarlaChallenge2020}. We find that we obtain a plateauing information gain with a 90\% confidence after only testing about 7.5\% of the scenario space, which is 23 \% less than what Latin Hypercube Sampling needs to explore to obtain the same quantity of information. %\todo{Nice!!}

To summarize, this work makes the following contributions:

\begin{itemize}
	\item A novel statistical representation of AV performance: a Bayesian Hierarchical Model, relying on quantitative evaluations of behavioral rules.
	\item A metric to quantify the relevance of a new scenario or set of scenarios: information gain.
	\item A scenario selection algorithm with near-optimality guarantees: the greedy algorithm for submodular optimization.
	\item A stopping criterion to contain the size of the scenario set: when the information gain is plateauing, with a given statistical confidence.
\end{itemize}

Section \ref{sec:lit} reviews related work in the scenario sampling literature. Section \ref{sec:method} details the steps used in the method described above, and section \ref{sec:exp} shows the results of the method in the CARLA experiment using simulated log data.

\section{Related Work} \label{sec:lit}

The problem of scenario sampling for AV testing and validation has attracted a lot of interest in recent years. We refer the reader to \cite{Riedmaier2020} for a more complete review of the field.

\subsection{AV Performance Representation}
Defining the ideal behavior for an AV is a difficult task because of the legal, ethical, safety objectives the system has to balance \cite{Censi2019}. The AV community has yet to agree upon a set of metrics to evaluate AV performance in a given scenario, however several metrics have been used. Active learning techniques indirectly exploit metrics for AV behavior as features that they learn preferences from \cite{Sadigh2017}.
Refs. \cite{Maierhofer2020, Xiao2021} propose sets of rules using temporal logic that AVs should strive to follow, such as minimizing collisions, staying within lane, and driving below the speed limit. Ref. \cite{Mehdipour2019} associates quantitative violation metrics with these logical statements. These works provide quantiative metrics for evaluating AV behavior across activities and ODDs. We want to evaluate an AV's metric violation rate across scenarios efficiently to either prompt a redesign if the aggregated violation is too high, or to validate the AV if it is at a later stage of development.

\subsection{Metrics for Value of a Scenario Set and Stopping Criteria}
To structure the scenario space, we take advantage of the geometric representation that uses features, or building blocks, as coordinates of a scenario vector \cite{Kruber2018a, DeGelder2020}. This modular view captures the combinatorial size of the scenario space. Using a finer feature representation increases the probability of having interesting scenarios, however it also increases the number of scenarios to be sampled. Similarity metrics and clustering techniques can be used to downsample the scenario space \cite{Kruber2018a}, however the number of scenarios still grows rapidly with the number of scenario features. Statistical hypothesis testing has been used to tackle the notion of completeness, or lack of redundancy, of a scenario set to provide a stopping criterion, but the ability to compute these metrics relies on major assumptions on the frequency of such scenarios in the real settings \cite{DeGelder2019, Hauer2019}.

\subsection{Scenario Sampling}
Some well-established sampling techniques aim at covering the space of possible scenarios to estimate AV performance. Many disciplines leverage Latin Hypercube Sampling (LHS) and Orthogonal Arrays to test the reliability of systems \cite{Olsson2003, Morris2008}. However, these techniques usually require independence between the scenario features they sample from. Ref. \cite{Zhao2017b} uses importance sampling to test AVs, but looks specifically for rare events. None of these techniques leverage the information obtained about the system in other ODDs to inform future sampling.

Other more recently developed techniques, such as adaptive stress-testing, reuse the information obtained in previous scenarios to choose the next ones, but they are usually focused on a specific functionality and aim at unveiling critical situations, rather than establishing an overall performance rate across ODDs \cite{Gangopadhyay2019,Mullins2017, Corso2019, Ponn2019b}. Machine learning techniques increase the efficiency of the sampling, especially for subsystem development \cite{Zhang2018, Joshi2010}, however they lack transparency and guarantees, in addition to adding difficulty in parameter tuning \cite{Langner2018}.

The goal of our method is to provide a scenario set to establish AV performance across ODDs efficiently, by leveraging information about the system obtained in previous scenarios.

\section{Method} \label{sec:method}
We first offer some background on Bayesian Hierarchical Models and how they can be constructed using ODD frameworks. We then demonstrate how this representation can be used to create a metric capturing information brought by a new scenario, how to determine a scenario set size, and how to sample scenarios efficiently.

\subsection{Plane}
Let $\mathcal{V}$ be the scenario space to sample from. A scenario $ S \in \mathcal{V}$ contains information about what is happening in the environment, outside of the system under test. We assume that we can characterize $ S $ as an element of $\mathbb{R}^n$, where $n$ is the number of features in the environment. These features can be based on scenario ontologies \cite{Bagschik2017}, or industry proposals for ODD definitions \cite{AutomatedVehicleSafetyConsortium2020}. In an additional and more bounded application, these coordinates could correspond to the parameters of a specific logical scenario. The scenarios considered can be generated in simulation or in real-life settings. Assuming $n$ features to describe a scenario such as cloud density, presence of a construction site, or number of dynamic agents, and $k$ potential values for each feature, $ |\mathcal{V}| = k^n$. For 50 features that could each take 6 values, this represents about $10^{10}$ scenarios. Although simulation allows testing an AV for the fraction of the cost of a track test, generating such a high number of scenarios still requires enormous computational capabilities.

We now need to define AV performance on any scenario. The Rulebooks framework provides rules, with associated violation metrics, to evaluate AV performance in a scenario according to an explicit behavior specification \cite{Collin2020a}. Indeed, AV performance is influenced by many different notions, such as traffic laws, courteous driving manners, or ride comfort \cite{Censi2019}. Even though our experiments in Section~\ref{sec:exp} use number of collisions, our method is compatible with the use of any performance metric. Once a specific rule has been translated into a formal logic statement, any trajectory in any scenario can be evaluated and given a violation score \cite{Mehdipour2019}.

We define as $X_\mathcal{V} = \{X_i, i \in \mathcal{V}\}$ the performance of the AV in the scenario space. $X_\mathcal{V}$ is a random variable and can be observed in scenarios. Our goal is to sample scenarios which will provide the most information about the distribution of $X_\mathcal{V}$. Thus, we need to establish some structure behind the distribution of $X_\mathcal{V}$. Indeed, if performance were completely chaotic, and there were no relation between rule violations in two different scenarios, none of the learnings from one scenario would be generalizable to another scenario, and we would need to test in every single scenario. Studies suggest that overall human driving performance is well-structured \cite{Bin-Nun2020}.

\begin{figure}[b]
	\centerline{\includegraphics[scale=0.3]{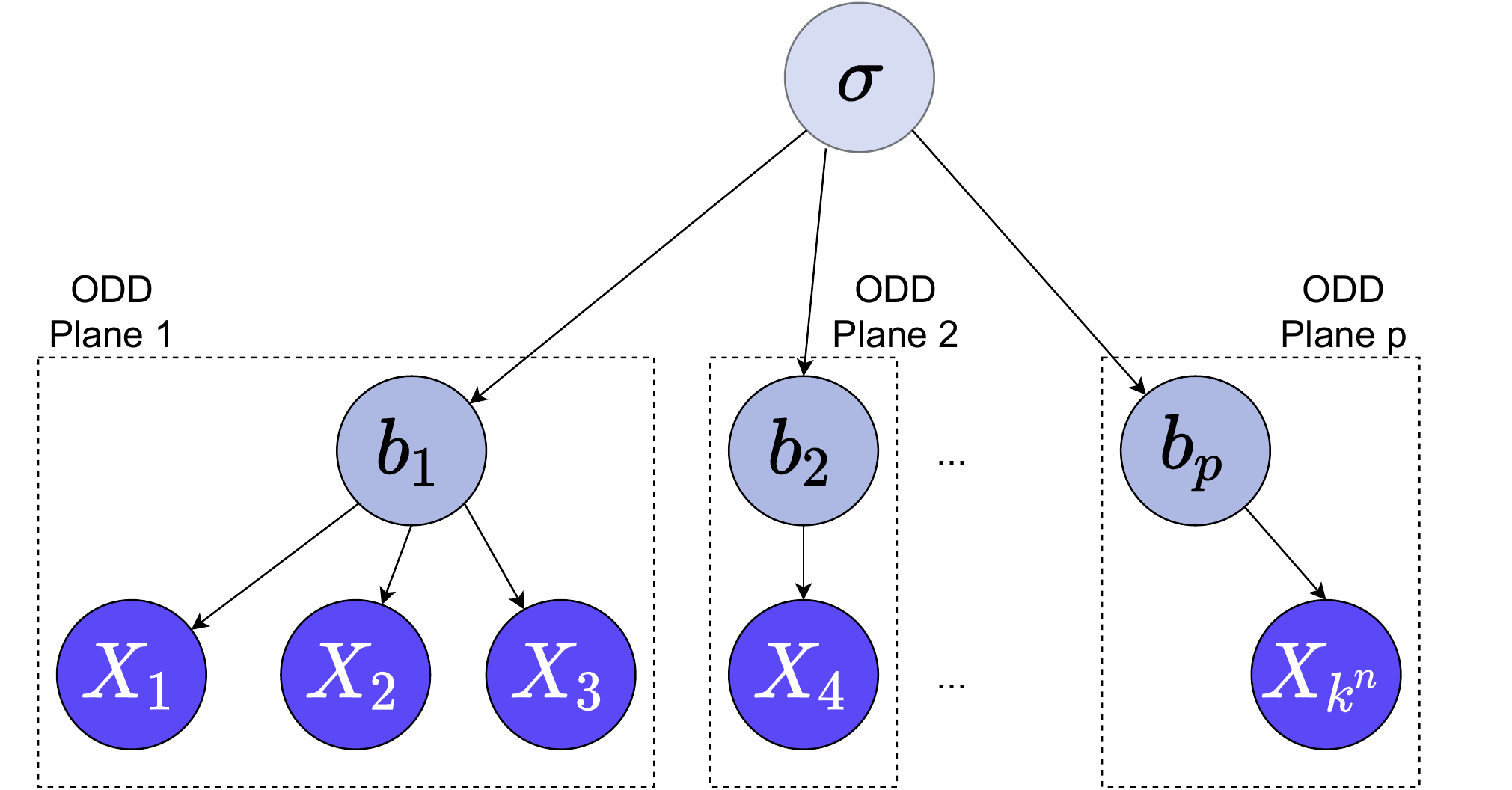}}
	\caption{Representation of the Bayesian Hierarchical Model. Light blue nodes represent unobserved variables, and purple nodes are the random variables of interest.}
	\label{fig:BayesianModel}
\end{figure}

We assume that a Bayesian Hierarchical Model can be fitted to observable AV performance (Fig.~\ref{fig:BayesianModel}). Because parameters are sampled from probability distributions, we use lower case letters to designate both the random variables and their realizations for the parameters and hyperparameter. %This representation uses observations in multiple ODD planes to infer overall AV performance. 
We define a hyperplane of the scenario space as the set of scenarios for which one feature, such as town of operation is fixed.
We therefore reformulate our assumption in the following manner: for $p$ hyperplanes of the scenario space, we assume that there exists $b_1, ..., b_p$ and $\sigma$ such that:
\begin{align*}
	X_i| b_p, \sigma & \sim P(x_i | b_p, \sigma) \\
	b_p| \sigma      & \sim P(b_p|\sigma)        \\
	\sigma           & \sim P(\sigma)
\end{align*}
where $\sigma$ is the hyperparameter with hyperprior $P(\sigma)$, $b_1, ..., b_p$ are generated from a population with distribution governed by the hyperparameter $\sigma$, and $P(x_i | b_p, \sigma)$ is the likelihood of AV performance, with $P(b_p,\sigma)$ as its prior distribution.
Choosing adequate priors and hyperpriors are the part of the model fitting process, and are out of scope for this paper \cite{Gelman2020}.

The interpretation of such a model is the following; AV performance is assumed to follow a distribution of parameter $b_p$ in the hyperplane $p$.
When testing in scenarios that take place in this town $p$, we effectively obtain samples $x_i$'s from this distribution. In a different town k, the shape of the distribution of AV performance will be the same, but the parameter $b_k$ is probably different, as $b_k$, $b_p$ are themselves sampled from $P(\sigma)$. %We show an example of Bayesian Hierarchical Model based on simulation data in . 
For example, AV's number of vehicle clearance violation over a scenario could follow a lognormal distribution, but the average could be higher in a dense town with more traffic than in a town with larger roads and fewer vehicles.

The quality of the scenario set determined by our method depends on the quality of the fit of the Bayesian Hierarchical Model to AV performance. A poor fit of parameters and hyperparameters would lead to a poor estimation of the information gain and the algorithm would therefore optimize for the wrong value. However, methods exist to ensure the quality of fit of a Bayesian Hierarchical Model, such as posterior predictive checks \cite{Gelman2020}.

\subsection{Sample}
This section focuses on how to leverage this AV performance representation for sampling.
We first use the following entropy definition to represent the uncertainty around the hyperparameter $\sigma$:
\begin{align*}
	H(\sigma) = -\sum_{\sigma} P(\sigma) \log P(\sigma)
\end{align*}

If $\mathcal{A}$ is the set of scenarios we observe, the information gain about $\sigma$ from observing performance in $\mathcal{A}$ is defined as
\begin{align*}
	I(\sigma;X_{\mathcal{A}}) = H(\sigma) - H(\sigma|X_{\mathcal{A}})
\end{align*}

There are $\binom{k^n}{a}$ scenario sets of size $a$ in this scenario space, with $a \leq k^n$.
The goal is to examine these options and choose the set that maximizes information gain before having to generate the scenarios in a simulation engine or in closed course testing. The optimization problem is therefore defined as:

\begin{align} \label{eq:objective}
	\max_{\mathcal{A} \subseteq V} f(\mathcal{A}) \quad s.t. \quad |\mathcal{A}| \leq C
\end{align}
where where $C$ is the scenario budget, and $f$ could be either $I(\sigma;X_{\mathcal{A}})$ for an overall AV assessment across the entire ODD, or $I(b_p;X_{\mathcal{A}})$ for an assessment of AV performance in one hyperplane, such as a specific town. $C$ is the equivalent of our stopping criterion. The conditional entropy $H(\sigma|X_{\mathcal{A}})$ can only be approximated up to a certain value with a given confidence interval \cite{Krause2012}. We propose to stop the optimization when the approximated information gain stops increasing in a statistically significant manner, which decides $C$.

This optimization problem is NP-hard \cite{Krause2012}, even in a simple setting where all scenarios have the same cost.
Nonetheless, both our potential objective functions are monotone \textit{submodular} \cite{Krause2014}. A function $f:2^V \to \mathbb{R}$ is submodular if and only if for every $A \subseteq B \subseteq V$ and $e \in V$,
\[ f(A \cup \{e\}) - f(A) \geq f(B \cup \{e\}) - f(B) \]

\begin{figure}[b]
	\centerline{\includegraphics[scale=0.3]{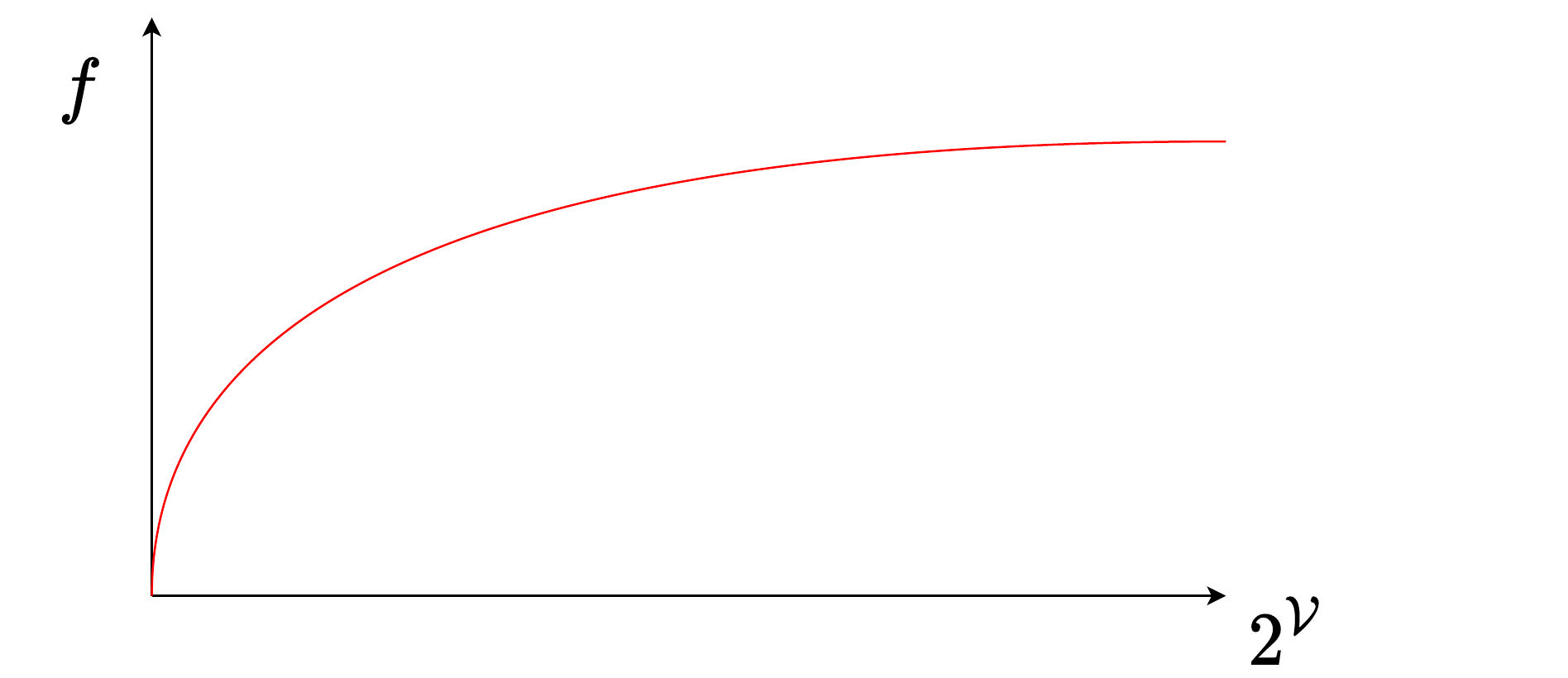}}
	\caption{Shape of a generic monotone submodular function. After more and more discrete elements are added to the set over which the function is defined, the function increases by a smaller and smaller amount.}
	\label{fig:submodular}
\end{figure}

This means that adding a scenario to an already large set will bring a lower information gain than adding a scenario to a smaller test set. This reflects our intuition that adding more and more scenarios to a set will yield diminishing returns, albeit positive returns (Fig~\ref{fig:submodular}).

The submodularity appears here because AV performances in different scenarios are conditionally independent given the hyperparameter $\sigma$. In general, the information gain about a random variable obtained from revealing another dependent random variable is not submodular. However, for a fixed AV stack, it is plausible that there exists a hyperplane in the scenario space that makes AV performance in one scenario conditionally independent from AV performance in another scenario given that hyperplane. For example, in section \ref{sec:exp}, within the same town of operation, the number of collisions in a scenario is independent from the number of collisions in another scenario given the AV's average number of collisions per route in that town. This means that, in our example, we assume that the probability distribution of the number of collisions is fully determined by the town of operation.

As long as this conditional independence property holds, the information gain about $\sigma$ upon revelation of $X_{\mathcal{A}}$ is submodular, as proven in~\cite{Krause2012}.

Since $f$ is submodular, we can use the greedy algorithm \cite{Nemhauser1978} to provide a near-optimal solution to equation (\ref{eq:objective}). Starting with an empty set of scenarios, this heuristic selects the next scenario with the highest information gain at each iteration:
\[\mathcal{A}_i = \mathcal{A}_{i-1} \cup \{ \argmax_e f(\mathcal{A}_{i-1} \cup \{e\}) - f(\mathcal{A}_{i-1}) \} \]
Since this selection problem is NP-hard, there is no algorithm that will find the solution in polynomial time. However, Ref.~\cite{Nemhauser1978} proves that the greedy algorithm finds a solution within $1-\frac{1}{e}\approx 0.63$ of the optimal value. Concretely, if the maximum information gain for $C$ scenarios is $1$, we are guaranteed to obtain a scenario set of size $C$ yielding at least $0.63$ information gain.

\section{Experiment and Results} \label{sec:exp}
We now demonstrate the use of this method with AV logs obtained from a CARLA simulation \cite{Dosovitskiy2017}.
\subsection{Data generation}
In this experiment, we use a model of an AV, available in CARLA, which is also used to simulate background traffic in simulations. The vehicle tries to follow routes in 6 different towns, with a changing number of traffic participants, for a total of 132 scenarios. These 3 features are the coordinates of the scenarios, shown in Table~\ref{tab:ScenarioDef}.

\begin{table}[b]
	\caption{Scenario Definitions}
	\begin{center}
		\begin{tabular}{|c|c|c|c|}
			%\hline
			%\textbf{Table}&\multicolumn{3}{|c|}{\textbf{Table Column Head}} \\
			%\cline{2-4} 
			\hline
			\textbf{Scenario ID} & \textbf{Number of traffic} & \textbf{Town} & \textbf{Route ID} \\
			                     & \textbf{participants}      &               &                   \\
			\hline
			1                    & 10                         & Town 02       & 1                 \\
			\hline
			2                    & 10                         & Town 02       & 2                 \\
			\hline
			\multicolumn{4}{|c|}{...}                                                             \\
			\hline
			132                  & 150                        & Town 06       & 49                \\
			%copy& More table copy$^{\mathrm{a}}$& &  \\
			\hline
			%\multicolumn{4}{l}{$^{\mathrm{a}}$Sample of a Table footnote.}
		\end{tabular}
		\label{tab:ScenarioDef}
	\end{center}
\end{table}

\begin{figure}[bthp]
	\vspace{0.2cm}
	\centering
	\begin{subfigure}[b]{0.49\columnwidth}
		\centering
		\includegraphics[width=\textwidth]{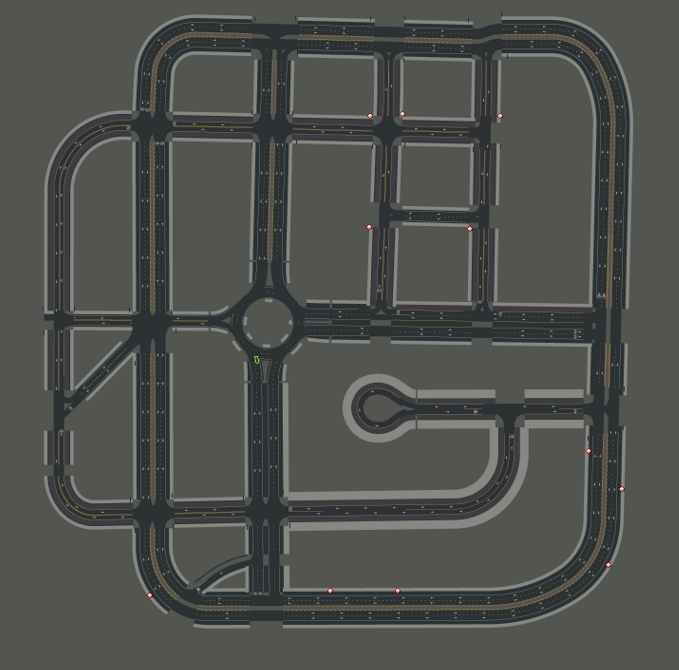}
		\caption{Town 03}
		\label{fig:y equals x}
	\end{subfigure}%
	\hfill
	\begin{subfigure}[b]{0.49\columnwidth}
		\centering
		\includegraphics[width=\textwidth]{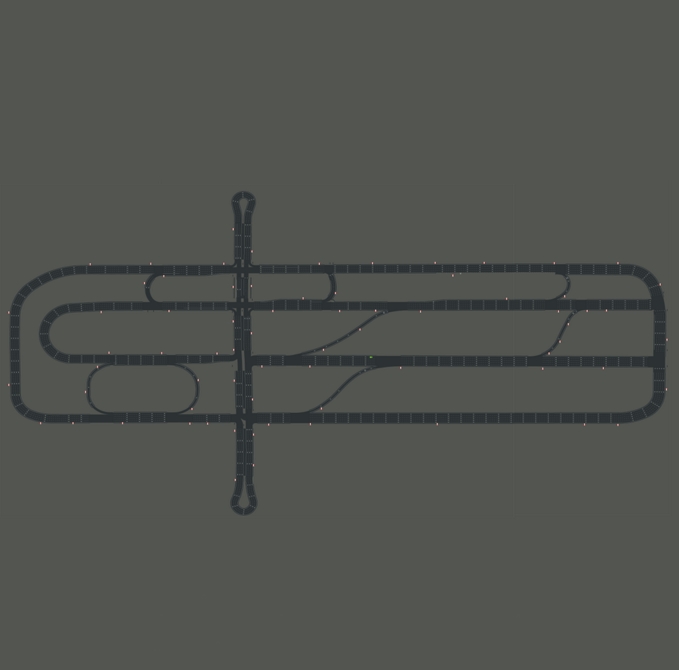}
		\caption{Town 06}
		\label{fig:three sin x}
	\end{subfigure}
	\caption{Maps of two different towns used for the scenarios.}
	\label{fig:two_maps}
\end{figure}

Fig.~\ref{fig:two_maps} shows maps for two of the towns considered; Town 03 has a lot more intersection and and roundabout, whereas Town 06 has long stretches of straight roads. Because of their large differences, the towns used in this experiment effectively represent different ODDs.

We record the number of collisions the AV was involved in in each scenario, and choose this as a performance metric $X_\mathcal{V}$.

\begin{figure}[b]
	\centerline{\includegraphics[scale=0.4]{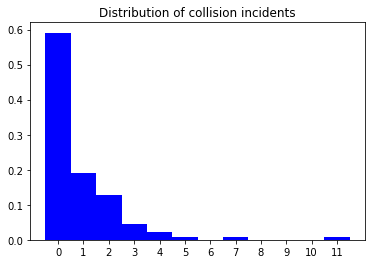}}
	\caption{Distribution of the number of collisions per scenario.}
	\label{fig:Collisions}
\end{figure}

The Poisson distribution is commonly used to express the probability of a number of events happening independently over fixed intervals \cite{Bertsekas2000}. Fig.~\ref{fig:Collisions} shows the probability distribution of $X_\mathcal{V}$ in all scenarios, which aligns well with the Poisson assumption. It is worth noting that the used AV implementation encounters a significant number of collisions per scenario because it does not have a perception module, as it is normally only used for background traffic, and only consumes map and route information. It therefore does not react to dynamic objects in the scene and would be expected to perform poorly. However, our method would be equally applicable to more sophisticated and better performing AVs; in fact, since it assesses emergent behavior, it is independent of the specific AV implementation.

\subsection{Bayesian Hierarchical Model}
\begin{figure}[bthp]
	\vspace{0.1cm}
	\centerline{\includegraphics[width=\columnwidth]{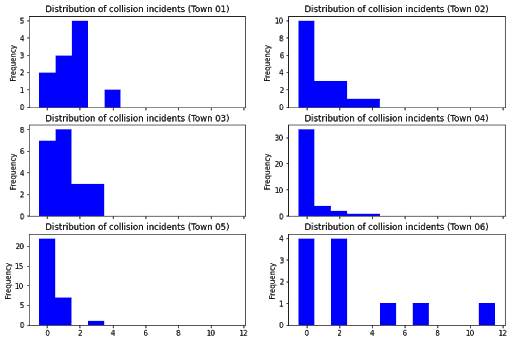}}
	\caption{Distributions of collisions along a route for the six different towns in CARLA.}
	\label{fig:CollisionPerTownDistributions}
\end{figure}

After data analysis, we find that the town is the scenario coordinate that has the biggest influence on the number of collisions per scenario (Fig. \ref{fig:CollisionPerTownDistributions}). This grouping can be inferred from a small amount of data that is already available, or decided based on expert knowledge.

\begin{figure}[b]
	\centering
	\begin{subfigure}[t]{0.49\columnwidth}
		\centering
		\includegraphics[width=\columnwidth, trim=0 0 442 0, clip]{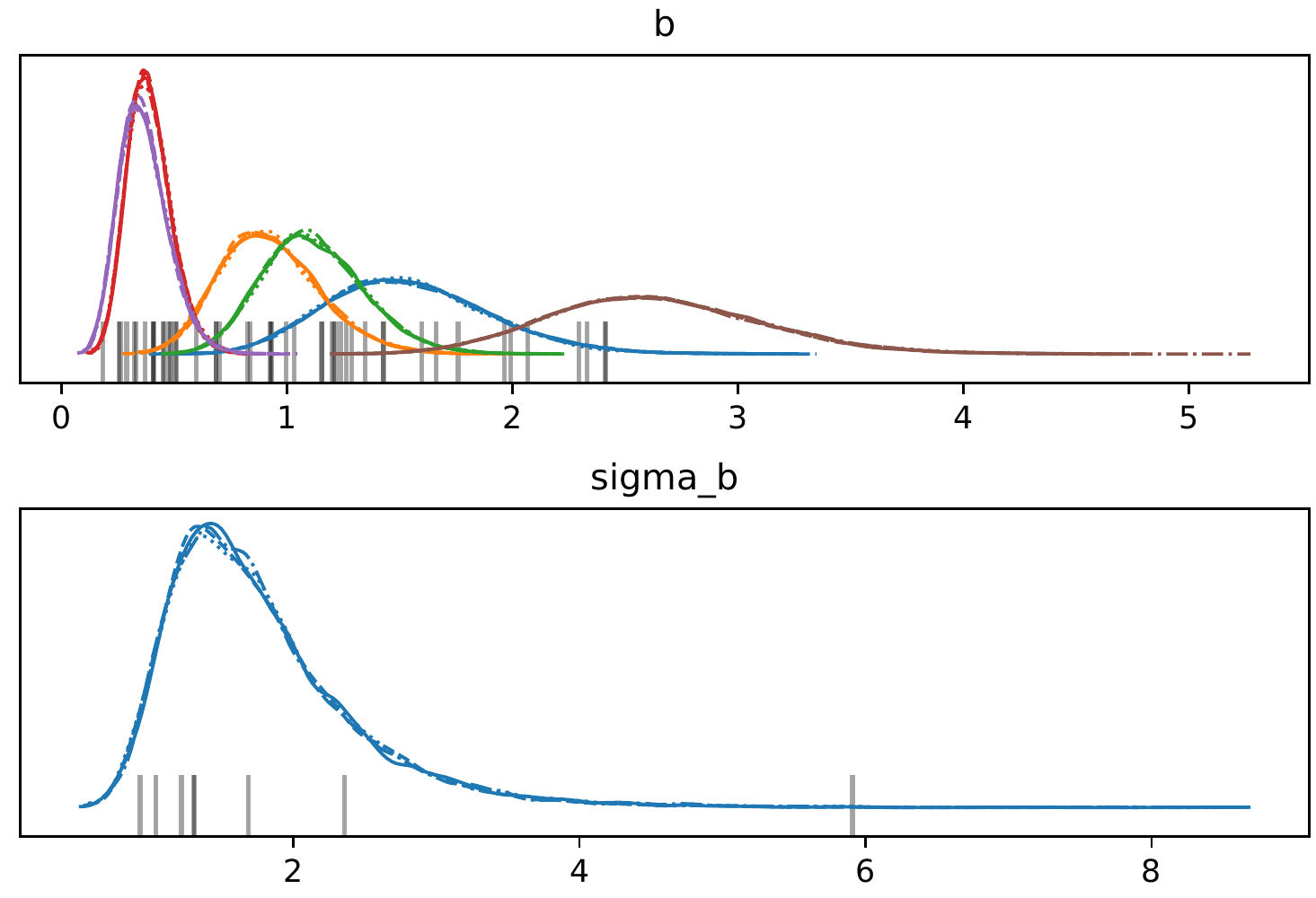}
		\caption{Posterior distribution of unobservable parameters}
		\label{fig:posteriors}
	\end{subfigure}%
	\begin{subfigure}[t]{0.49\columnwidth}
		\centering
		\includegraphics[width=\columnwidth, trim= 22 0 22 0, clip]{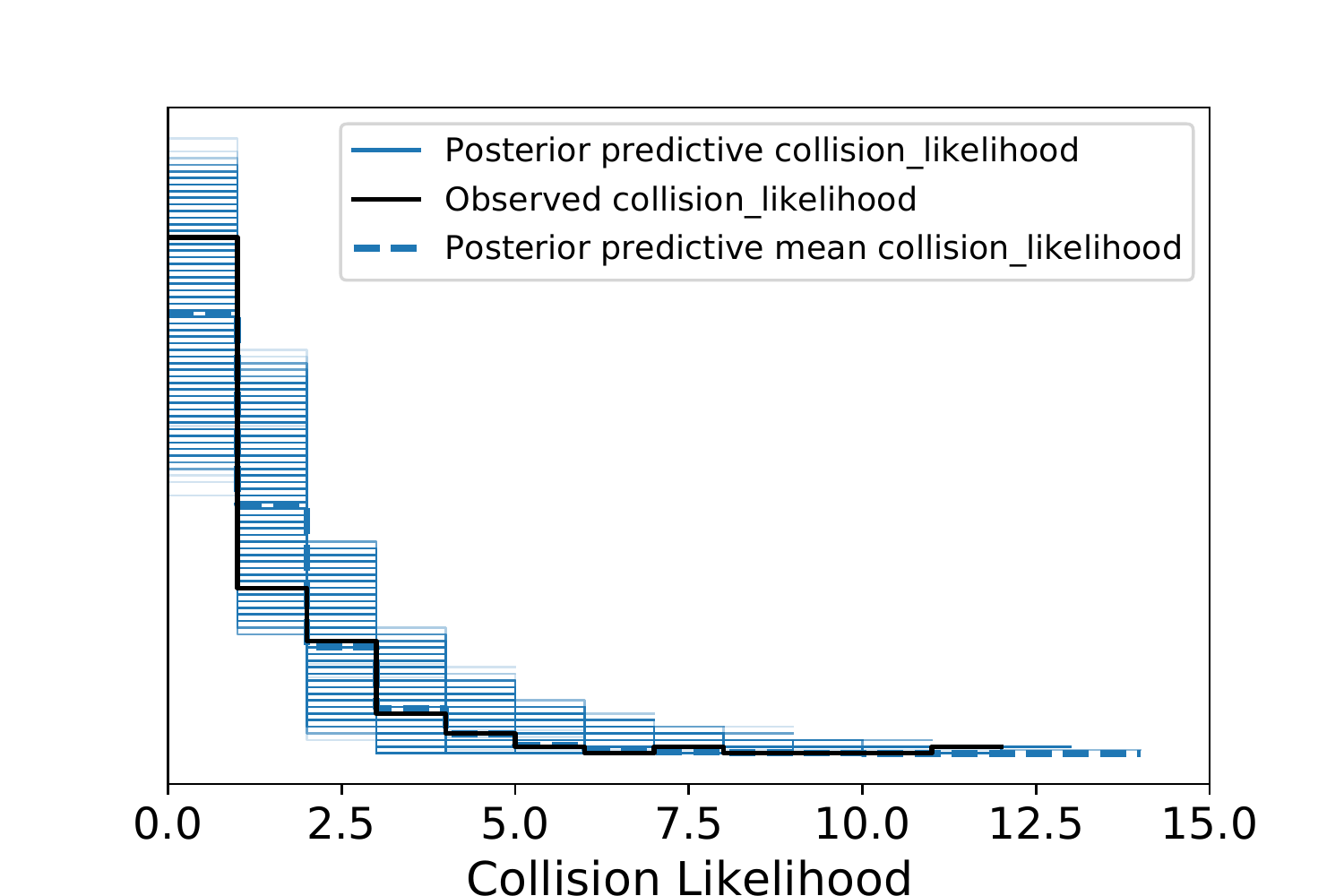}
		\caption{Posterior predictive fit.}
		\label{fig:posterior_predictive}
	\end{subfigure}
	\caption{Bayesian Hierarchical Model Results}
	\label{fig:Bayesian_results}
\end{figure}

We therefore slice the scenario space by town, and fit the following Bayesian Hierarchical Model:
\begin{align*}
	X_i    & \sim \text{Poisson}(b_p)       \\
	b_p    & \sim \text{HalfNormal}(\sigma) \\
	\sigma & \sim \text{HalfNormal}(5)      \\
\end{align*}
We chose the half-normal distribution for the parameters $b_p$ and hyperparameter $\sigma$ as a weakly informative priors. The goal of the Bayesian Hierarchical Model fit is to find the parameters for these half-normal distributions. We use PyMC3 \cite{Salvatier2016} to perform this fit. The results are shown in Fig.~\ref{fig:posteriors}. The resulting average number of collisions per scenario varies greatly, between less than 1 for some towns and more than 2.5 for others. A posterior predictive check shows that the average value of the graph-based re-generated samples of $X_\mathcal{V}$ aligns well with the observed values, except for $\{0,1\}$ (Fig.~\ref{fig:posterior_predictive}). This means that the fit is better for larger values of the random variable, however it is still producing values within the observed region for 0 and 1. %This Bayesian Hierarchical Model is selected for the next step of the method.

\subsection{Scenario selection}
We proceed to the optimization of the information gain on AV performance across ODDs, but the same method can be applied to choose a scenario set to maximize information gain for a specific town. The scenario sampling problem therefore becomes:
\begin{align} \label{optim}
	\max_{\mathcal{A} \subseteq V} I( \sigma;X_{\mathcal{A}}) \quad s.t. \quad |\mathcal{A}| \leq C
\end{align}
The Bayesian Hierarchical Model allows the computation of $P(\sigma|X)$, which is then used to compute the conditional entropy \cite{Krause2012} and the information gain of any new scenario. As long as a scenario can be placed in the graph, the added information it provides can be inferred.

Algorithm~\ref{alg:greedy}, adapted from \cite{Krause2012} with our stopping criterion, details the steps to solve (\ref{optim}). The conditional entropy $H(\sigma|X_{\mathcal{A}})$ is computed up to a 90\% confidence interval on an absolute error of 0.1. At each step, a scenario is added to the set $\mathcal{A}$ by greedily looking at each scenario one-by-one, computing the information gained from revealing performance in this scenario compared with the previously chosen scenarios, and taking the one that has the highest information gain for this step. This is why this algorithm is said to break ties arbitrarily \cite{Krause2014}. This procedure is repeated until the information gain stops growing (within 90 \% confidence).

\begin{algorithm}[tbhp]
	\caption{Greedy Algorithm}
	\label{alg:greedy}
	\begin{algorithmic}[1]
		\renewcommand{\algorithmicrequire}{\textbf{Input:}}
		\renewcommand{\algorithmicensure}{\textbf{Output:}}
		\REQUIRE $\mathcal{V}$ the list of all possible scenarios, $M$ an inference model
		\ENSURE $\mathcal{A}^*$ a near-optimal, finite, scenario selection
		\\ \textit{Initialization}
		\STATE $\mathcal{A} = \emptyset$, $I_0=H(\sigma)$, $I_{\text{diff}} = +\infty$, $I^*=0$, $k=1$
		\\ \textit{Add scenarios to the set until there is no statistically significant information gain}
		\WHILE{$I_{\text{diff}} > 0$}
		%\STATE $I = +\infty$
		\FOR{$S \in \mathcal{V} \setminus \mathcal{A}$}
		\STATE perform inference on $M$ to get $P(\sigma|X_{S})$
		\STATE I = $H(\sigma) - H(\sigma|X_{S})$
		\IF{$I > I^*$}
		\STATE $I^* = I$
		\STATE $S^* = S$
		\ENDIF
		\ENDFOR
		\STATE $\mathcal{A} = \mathcal{A} \cup S^*$
		\STATE $I_k = I^*$
		\STATE $I_{\text{diff}} = I_k - I_{k-1}$
		\STATE $k=k+1$
		\ENDWHILE
		\RETURN $\mathcal{A}$
	\end{algorithmic}
\end{algorithm}

\begin{figure}[t]
	\vspace{0.2cm}
	\centerline{\includegraphics[width=0.98\columnwidth]{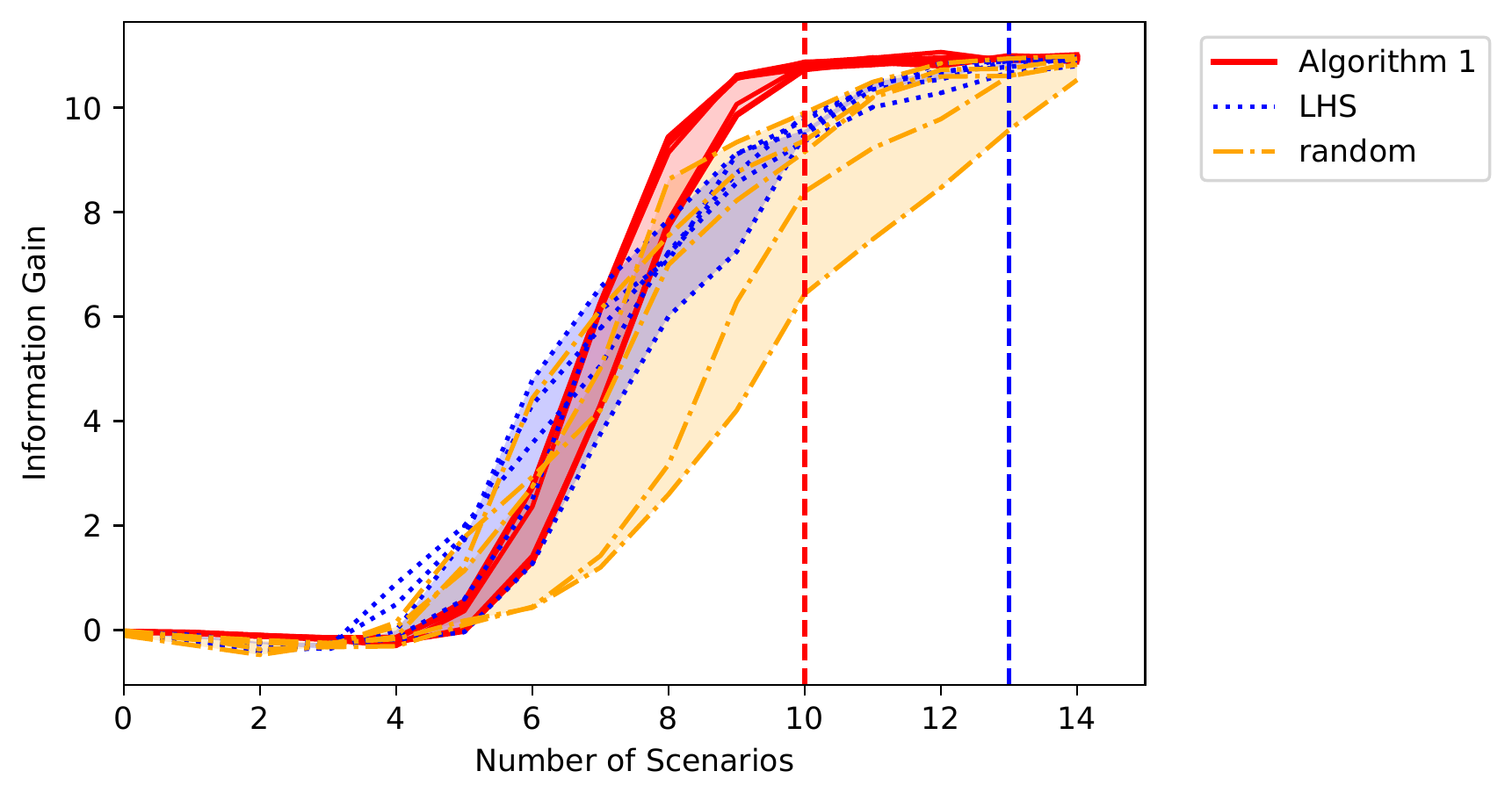}}
	\caption{Comparison of information gain and stopping criterion between our algorithm (red), Latin Hypercube Sampling (blue), and random (yellow). Each algorithm is run 5 times, and the colored surfaces show the ranges of results obtained. The vertical lines show our stopping criterion for our algorithm and for LHS for most runs.}
	\label{fig:final_info_gain}
\end{figure}

We compare the results of Algorithm~\ref{alg:greedy} with several runs of a Latin Hypercube Sampling implementation \cite{Minasny2006} and several random selections. The results are presented in Fig.~\ref{fig:final_info_gain}.

We first notice that all selections show the diminishing returns property, which confirms the submodularity property of the information gain, except for the first few scenarios. This is probably due to the approximation in computing this gain, which has a larger impact at the beginning for values close to zero. Additionally, using only a couple of observations might lead to a degenerated version of the model, which could explain the lack of submodularity for the first few observations. Because of this, LHS may sometimes perform better than Alg.~\ref{alg:greedy} with very few scenarios, but not on average.

Regarding the validation of our choice of metric to assess the relevance of a scenario set, we observe that the information gain captures the fact that a well establish method such as LHS indeed performs better on average than random selections. This metric allows the comparison of any scenario selection method, regardless of ODD or activity analyzed. It also conveys a sense of progress in the evaluation campaign, and is intuitive and transparent - when we stop receiving significant amounts of information, we can stop the evaluation.

The main result in Fig.~\ref{fig:final_info_gain} is the following: with all sampling methods plateauing at the same information gain level, our proposed algorithm always obtains this quantity with only 10 scenarios or less, or about 7.5 \% of the scenario space, whereas LHS needs 13 scenarios or more, or about an extra 2 \% of the scenario space, and random methods usually need to explore at least another 2 \% of the space. This experiment is performed at a small scale, but these differences could represent a gain of millions of scenarios in a large scale AV validation process. The chosen set consistently contains scenarios taking place in 4 different towns, but not all of them. Scenarios in towns 02 or 03 are not predicted to bring as much information as scenarios in towns 01, 04, 05, and 06. When the AV is actually tested through the chosen scenarios, it encounters mostly 0 or 1 collision but also is involved in 7 collisions in one scenario and in 11 collisions in another. Although the algorithm does not look for rare events, these scenarios at the tail of the distribution provide a lot of information about the system under test.

\section{Conclusion} \label{sec:conclusion}
In this paper, we frame the scenario selection problem for AV performance evaluation as a submodular optimization problem on a graph. In doing so, we also propose a novel statistical representation of AV performance, leveraging recent advances in quantitative AV behavior definition \cite{Censi2019}.
We are able to easily fit a Hierarchical Bayesian Model to a dataset generated by an open source AV implementation, suggesting that this method can be applied to any AV implementation. The main assumption of the method is that there is enough data to start with to construct a Bayesian Hierarchical Model, and that such a model fits well.

The main limitation of this method is therefore linked to performance metrics that would not be easily characterized by well-known distributions, such as multi modal distributions.
The starting dataset used to fit the model should also not be biased, for instance by only using data where the AV had more than 3 collisions, otherwise the inferred distribution and resulting scenario set will be biased as well.

Once the Bayesian Hierarchical Model is validated, the submodularity property ensues, and we show how to build a scenario set leveraging this property. Information gain can be used to compare scenario selection methods regardless of ODD or functionality examined. Future work involves running this method on a larger scale experiment as well as exploring other violation metrics with potentially different shapes of distributions.

\bibliographystyle{BibFiles/IEEEtranUrldate.bst}

\bibliography{PlaneAndSample}

\vspace{15cm}

\end{document}